\documentstyle[aclap]{article}

\title{Assigning Grammatical Relations with a Back-off Model} 

\author{Erika F. de Lima\\
GMD - German National Research Center
for Information Technology\\ 
Dolivostrasse 15\\ 
64293 Darmstadt, Germany\\{\tt delima@darmstadt.gmd.de}}

\newcounter{C}
\stepcounter{C}

\newenvironment{sentence}%
{ \medskip \noindent\begin{tabular}[t]{@{}l@{\hspace{1.1mm}}l@{\hspace{1.1mm}}l@{\hspace{1.1mm}}l@{\hspace{1.1mm}}l@{\hspace{1.1mm}}l@{\hspace{1.1mm}}l@{\hspace{1.1mm}}l@{\hspace{1.1mm}}l@{\hspace{1.1mm}}l@{\hspace{1.1mm}}l@{\hspace{1.1mm}}l@{\hspace{1.1mm}}l@{\hspace{1.1mm}}l@{\hspace{1.1mm}}l@{\hspace{1.1mm}}l}}%
{ \end{tabular} }

\newenvironment{starredsentence}%
{ \medskip \noindent\begin{tabular}[t]{@{}l@{\hspace{1.1mm}}l@{}l@{\hspace{1.1mm}}l@{\hspace{1.1mm}}l@{\hspace{1.1mm}}l@{\hspace{1.1mm}}l@{\hspace{1.1mm}}l@{\hspace{1.1mm}}l@{\hspace{1.1mm}}l@{\hspace{1.1mm}}l@{\hspace{1.1mm}}l@{\hspace{1.1mm}}l@{\hspace{1.1mm}}l@{\hspace{1.1mm}}l@{\hspace{1.1mm}}l}}%
{ \end{tabular} }

\newenvironment{continuation}%
{ \smallskip \noindent\begin{tabular}{@{}ll@{\hspace{1.1mm}}l@{\hspace{1.1mm}}l@{\hspace{1.1mm}}l@{\hspace{1.1mm}}l@{\hspace{1.1mm}}l@{\hspace{1.1mm}}l@{\hspace{1.1mm}}l@{\hspace{1.1mm}}l@{\hspace{1.1mm}}l@{\hspace{1.1mm}}l@{\hspace{1.1mm}}l@{\hspace{1.1mm}}l@{\hspace{1.1mm}}l@{\hspace{1.1mm}}l}}%
{ \end{tabular} }

\newenvironment{sentenceNarrow}%
{ \medskip \noindent\begin{tabular}[t]{@{}l@{\hspace{1.1mm}}l@{\hspace{0.8mm}}l@{\hspace{0.8mm}}l@{\hspace{0.8mm}}l@{\hspace{0.8mm}}l@{\hspace{0.8mm}}l@{\hspace{0.8mm}}l@{\hspace{0.8mm}}l@{\hspace{0.8mm}}l@{\hspace{0.8mm}}l@{\hspace{0.8mm}}l@{\hspace{0.8mm}}l@{\hspace{0.8mm}}l@{\hspace{0.8mm}}l@{\hspace{0.8mm}}l}}%
{ \end{tabular} }

\newenvironment{continuationNarrow}%
{ \smallskip \noindent\begin{tabular}{@{}ll@{\hspace{1.1mm}}l@{\hspace{0.8mm}}l@{\hspace{0.8mm}}l@{\hspace{0.8mm}}l@{\hspace{0.8mm}}l@{\hspace{0.8mm}}l@{\hspace{0.8mm}}l@{\hspace{0.8mm}}l@{\hspace{0.8mm}}l@{\hspace{0.8mm}}l@{\hspace{0.8mm}}l@{\hspace{0.8mm}}l@{\hspace{0.8mm}}l@{\hspace{0.8mm}}l}}%
{ \end{tabular} }

\begin{document}
\bibliographystyle{fullname}
\maketitle

\begin{abstract}
This paper presents a corpus-based method to assign grammatical
subject/object relations to ambiguous German constructs. It makes use
of an unsupervised learning procedure to collect training and test
data, and the back-off model to make assignment decisions.
\end{abstract}

\section{Introduction}

Assigning a parse structure to the German sentence (1) involves
addressing the fact that it is syntactically ambiguous:
 
\begin{sentence}
(\arabic{C})
& Eine hohe Inflationsrate & erwartet & die \"Okonomin.\\
& a high inflation rate & expects & the economist\\
&\makebox[0mm][l]{`The economist expects a high inflation rate.'}\\
\end{sentence}
\stepcounter{C}

\medskip

\noindent
In this sentence it must be determined which nominal phrase is the
subject of the verb.  The verb {\em erwarten} (`to expect') takes, in
one reading, a nominative NP as its subject and an accusative NP as
its object. The nominal phrases preceding and following the verb in
(1) are both ambiguous with respect to case; they may be nominative or
accusative. Further, both NPs agree in number with the verb, and since
in German any major constituent may be fronted in a verb-second
clause, both NPs may be the subject/object of the verb. In this
example, morpho-syntactical information is not sufficient to determine
that the nominal phrase [$_{NP}$~die~\"Okonomin] (`the~economist') is
the subject of the verb, and [$_{NP}$~Eine hohe~Inflationsrate]
(`a~high inflation rate') its object.

Determining the subject/object of an ambiguous construct such as (1)
with a knowledge-based approach requires (at least) a lexical
representation specifying the classes of entities which may serve as
arguments in the relation(s) denoted by each verb in the vocabulary,
as well as membership information with respect to these classes for
all entities denoted by nouns in the vocabulary. One problem with this
approach is that it is usually not available for a broad-coverage
system.

This paper proposes an approximation, similar to the empirical
approaches to PP attachment decision \cite{HR93,RR94,CoBro95}. These
make use of unambiguous examples provided by a treebank or a learning
procedure in order to train a model to decide the attachment of
ambiguous constructs. In the current setting, this approach involves
learning the classes of nouns occurring unambiguously as
subject/object of a verb in sample text, and using the classes thus
obtained to disambiguate ambiguous constructs.

Unambiguous examples are provided by sentences in which
morpho-syntactical information suffices to determine the subject and
object of the verb.  For instance in (2), the nominal phrase
[$_{NP}$~der \"Okonom] with a masculine head noun is unambiguously
nominative, identifying it as the subject of the verb.  In (3), both
NPs are ambiguous with respect to case; however, the nominal phrase
[$_{NP}$~Die \"Okonomen] with a plural head noun is the only one to
agree in number with the verb, identifying it as its subject.

\begin{sentence}
(\arabic{C})
& Eine hohe Inflationsrate & erwartet & der \"Okonom.\\
& a   high   inflation rate & expects & the economist \\
&\makebox[0mm][l]{`The economist expects a high inflation rate.'}\\
\end{sentence}
\stepcounter{C}

\begin{sentenceNarrow}
(\arabic{C})&
 Die \"Okonomen & erwarten & eine hohe Inflationsrate.\\
 & the economists & expect & a high inflation rate \\
& \makebox[0mm][l]{`The economists expect a high inflation rate.'}\\
\end{sentenceNarrow}
\stepcounter{C}
\medskip

\noindent
This paper describes a procedure to determine the subject and object
in ambiguous German constructs automatically.  It is based on shallow
parsing techniques employed to collect training and test data from
(un)ambiguous examples in a text corpus, and the back-off model to
determine which NP in a morpho-syntactically ambiguous construct is
the subject/object of the verb, based on the evidence provided by the
collected training data.

\section{Collecting Training and Test Data}
\label{sec-collect}

\noindent
Shallow parsing techniques are used to collect training and test data
from a text corpus.  The corpus is tokenized, morphologically
analyzed, lemmatized, and parsed using a standard CFG parser with a
hand-written grammar to identify clauses containing a finite verb
taking a nominative NP as its subject and an accusative NP as its
object.

Constructs covered by the grammar include verb-second and verb-final
clauses.  Each clause is segmented into phrase-like constituents,
including nominative (NC), prepositional (PC), and verbal (VC)
constituents. Their definition is non-standard; for instance, all
prepositional phrases, whether complement or not, are left unattached.
As an example, the shallow parse structure for the sentence in
(\arabic{C}) is shown in (\arabic{C}$'$) below.

\begin{sentenceNarrow}
(\arabic{C}) 
 & Die Gesellschaft & erwartet & in diesem Jahr \\
 & the society      & expects & in this year \\
\end{sentenceNarrow}
\begin{continuationNarrow}
 \makebox[1.7mm] & in S\"udostasien &einen Umsatz \\
 & in southeast Asia & a turnover \\
\end{continuationNarrow}
\begin{continuationNarrow}
 \makebox[1.7mm] & von 125 Millionen DM.\\
                 & from 125 million DM\\[1mm]
&\makebox[0mm][l]{`The society expects this year in southeast Asia}\\
 &\makebox[0mm][l]{ a turnover of 125 million DM.'}\\
\end{continuationNarrow}

\begin{tabbing}
\=\hspace{0.9cm}\=\hspace{0.5cm}\= \kill
 \> (\arabic{C}$'$) \>  [$_{S}$ \> [$_{NC_{3,s,\{nom,acc\}}}$ Die Gesellschaft] \\
 \>  \>      \> [$_{VC_{3,s}}$ erwartet] \\
 \>  \>      \> [$_{PC}$ in diesem Jahr] \\
 \>  \>      \> [$_{PC}$ in S\"udostasien] \\
 \>  \>      \> [$_{NC_{3,s,acc}}$ einen Umsatz] \\
 \>  \>      \> [$_{PC}$ von 125 Millionen DM] \\
\>  \>      ] 
\end{tabbing}

\noindent
Nominal and verbal constituents display person and number information;
nominal constituents also display case information.  For instance in
the structure above, {\em 3} denotes third person, {\em s} denotes singular
number, {\em nom} and {\em acc} denote nominative and accusative case,
respectively. The set $\{nom, acc\}$ indicates that the first nominal
constituent in the structure is ambiguous with respect to case; it may
be nominative or accusative.

Test and training tuples are obtained from shallow structures
containing a verbal constituent and two nominative/accusative nominal
constituents. Note that no subcategorization information is used; it
suffices for a verb to occur in a clause with two
nominative/accusative NCs for it to be considered testing/training
data.

Training data consists of tuples $(n_1, v, n_2, x)$, where $v$ is a
verb, $n_1$ and $n_2$ are nouns, and $x~\in~\{1,0\}$ indicates whether
$n_1$ is the subject of the verb. Test data consists of ambiguous
tuples $(n_1, v, n_2)$ for which it cannot be established which noun
is the subject/object of the verb based on morpho-syntactical
information alone.

The set of training and test tuples for a given corpus is obtained as
follows.  For each shallow structure $s$ in the corpus containing one
verbal and two nominative/accusative nominal constituents, let $n_1,
v, n_2$ be such that $v$ is the main verb in $s$, and $n_1$ and $n_2$
are the heads of the nominative/accusative NCs in $s$ such that $n_1$
precedes $n_2$ in $s$. In the rules below, $i,j \in \{1,2\}, j \neq
i$, and $g(i) = 1$ if $i = 1$, and 0 otherwise. Note that the last
element in a training tuple indicates whether the first NC in the
structure is the subject of the verb (1 if so, 0 otherwise).

\begin{trivlist}
\item {\bf Case Nominative Rule.} If $n_i$ is masculine, and the NC
headed by $n_i$ is unambiguously nominative\footnote{Only NCs with a
masculine head noun may be unambiguous with respect to
nominative/accusative case in German.}, then $(n_1, v, n_2, g(i))$ is
a training tuple,

\item {\bf Case Accusative Rule.} If $n_i$ is masculine, and the NC
headed by $n_i$ is unambiguously accusative, then $(n_1, v, n_2,
g(j))$ is a training tuple,

\item {\bf Agreement Rule.} If $n_i$ but not $n_j$ agrees with $v$ in
person and number, then $(n_1, v, n_2, g(i))$ is a training tuple,

\item {\bf Heuristic Rule.} If the shallow structure consists of a
verb-second clause with an adverbial in the first position, or of a
verb-final clause introduced by a conjunction or a complementizer,
then $(n_1, v, n_2,1)$ is a training tuple (see below for examples),

\item {\bf Default Rule.} $(n_1, v, n_2)$ is a test triple.
\end{trivlist}

For instance, the training tuple {\em (Gesellschaft, erwarten, Umsatz,
1)} (`society, expect, turnover') is obtained from the structure
(\arabic{C}$'$) above with the Case Accusative Rule, since the NC
headed by the masculine noun {\em Umsatz} (`turnover') is
unambiguously accusative and hence the object of the verb.  The
training tuple {\em (Inflationsrate, erwarten, \"Okonom, 0)}
(`inflation rate, expect, economist') and {\em (\"Okonom, erwarten,
Inflationsrate, 1)} (`economist, expect, inflation rate') are obtained
from sentences (2) and (3) with the Case Nominative and Agreement
Rules, respectively, and the test tuple {\em (Inflationsrate,
erwarten, \"Okonomin)} (`inflation rate, expect, economist') from the
ambiguous sentence in (1) by the Default Rule. \stepcounter{C}

The Heuristic Rule is based on the observation that in the constructs
stipulated by the rule, although the object may potentially precede
the subject of the verb, this does not (usually) occur in written
text.  (\arabic{C}) and (6) are sentences to which this rule applies.

\begin{sentence}
(\arabic{C})
& In diesem Jahr & erwartet & die \"Okonomin \\
& in this year &  expects & the economist\\
\end{sentence}
\begin{continuation}
 \makebox[1.7mm]  & eine hohe Inflationsrate.\\
  & a high inflation rate \\[1mm]
& \makebox[0mm][l]{`This year the economist expects a high }\\
&\makebox[0mm][l]{ inflation rate.''}\\
\end{continuation}

\stepcounter{C}
\begin{sentence}
(\arabic{C}) & Weil & die \"Okonomin & eine hohe Inflationsrate\\
             & because & the economist & a high inflation rate \\
\end{sentence}
\begin{continuation}
 \makebox[1.7mm]  & erwartet, \ldots \\
                  & expects \\[1mm]
& \makebox[0mm][l]{`Because the economist expects a high inflation}\\
&\makebox[0mm][l]{ rate, \ldots '}\\
\end{continuation}
\stepcounter{C}
\medskip

\noindent 
Note that the Heuristic Rule does not apply to verb-final clauses
introduced by a relative or interrogative item, such as in (7):

\begin{sentenceNarrow}
(\arabic{C}) 
& Die Rate, & die & die \"Okonomin  & erwartet, \ldots\\
& the rate & which & the economist & expects, \ldots\\
\end{sentenceNarrow}

\section{Testing}

The testing algorithm makes use of the back-off model \cite{Ka87} in
order to determine the subject/object in an ambiguous test tuple.  The
model, developed within the context of speech recognition, consists of
a recursive procedure to estimate $n$-gram probabilities from sparse
data. Its generality makes it applicable to other areas; the method
has been used, for instance, to solve prepositional phrase attachment
in \cite{CoBro95}.

\subsection{Katz's back-off model}

Let $w_1^n$ denote the $n$-gram $w_1, \ldots, w_n$, and $f(w_1^n)$
denote the number of times it occurred in a sample text. The back-off
estimate computes the probability of a word given the $n-1$ preceding
words. It is defined recursively as follows. (In the formulae below,
$\alpha(w_1^{n-1})$ is a normalizing factor and $d_r$ a discount
coefficient. See \cite{Ka87} for a detailed account of the model.)


$$
P_{bo}(w_n|w_1^{n-1}) \!= \!\left\{
\!\!\begin{array}{l}
\!\!\tilde{P}(w_n|w_1^{n-1}), \mbox{ if $\tilde{P}(w_n|w_1^{n-1})>0$}\\
\!\!\alpha(w_1^{n-1})P_{bo}(w_n|w_2^{n-1}), \mbox{ otherwise,}
\end{array}
\right.
$$

\noindent
where $\tilde{P}(w_n|w_1^{n-1})$ is defined as follows:\nopagebreak

$$
\tilde{P}(w_n|w_1^{n-1}) = \left\{
\begin{array}{ll}
d_{f(w_1^n)}\frac{f(w_1^n)}{f(w_1^{n-1})}, & \mbox{if $f(w_1^{n-1}) \neq 0$}\\
0, & \mbox{otherwise.}
\end{array}
\right.
$$

\subsection{The Revised Model}
\label{rev-sec}

In the current context, instead of estimating the probability of a
word given the $n-1$ preceding words, we estimate the probability that
the first noun $n_1$ in a test triple $(n_1, v, n_2)$ is the
subject of the verb $v$, i.e., $P(S=1|N_1=n_1,V=v,N_2=n_2)$ where $S$
is an indicator random variable ($S=1$ if the first noun in the triple
is the subject of the verb, 0 otherwise).

In the estimate $
P_{bo}(w_n|w_1^{n-1})$ only one relation---the
precedence relation---is relevant to the problem; in the current
setting, one would like to make use of two implicit relations in the
training tuple---subject and object---in order to produce an estimate
for $P(1|n_1,v,n_2)$. The model below is similar to that in
\cite{CoBro95}.

Let $\cal L$ be the set of lemmata occurring in the training triples
obtained from a sample text, and let $c(n_1,v,n_2,x)$ denote the
frequency count obtained for the training tuple $(n_1,v,n_2,x)$ ($x
\in \{0,1\}$).  We define the count $f_{so}(n_1,v,n_2) =
c(n_1,v,n_2,1)+ c(n_2,v,n_1,0)$ of $n_1$ as the subject and $n_2$ as
the object of $v$. Further, we define the count $f_s(n_1,v) =
\sum_{n_2 \in {\cal L}} f_{so}(n_1,v,n_2)$ of $n_1$ as the
subject of $v$ with any object, and analogously, the count $f_o(n_1,
v)$ of $n_1$ as the object of $v$ with any subject. Further, we define
the counts $f_s(v) = \sum_{n_1,n_2 \in {\cal L}}
c(n_1,v,n_2,1)$ and $f_o(v) = \sum_{n_1,n_2 \in {\cal L}}
c(n_1,v,n_2,0)$. The estimate $P_i(1|n_1,v,n_2)$ $(0 \le i \le 3)$ is
defined recursively as follows:

$$
\begin{array}{l}

P_0(1|n_1,v,n_2) = 1.0

\\

P_i(1|n_1,v,n_2) = \left\{
\begin{array}{l}
\frac{c_i(n_1,v,n_2)}{t_i(n_1,v,n_2)}, \mbox{ if $t_i(n_1,v,n_2)> 0$}\\
P_{(i-1)}(1|n_1,v,n_2), \mbox{ otherwise,}
\end{array}
\right.

\end{array}
$$
\noindent
where the counts $c_i(n_1,v,n_2)$, and $t_i(n_1,v,n_2)$ are
defined as follows:

$$
c_i(n_1,v,n_2) = \left\{
\begin{array}{ll}
f_{so}(n_1,v,n_2), & \mbox{if $i = 3$}\\
f_s(n_1,v) + f_o(n_2,v), & \mbox{if $i = 2$}\\
f_s(v), & \mbox{if $i = 1$}\\
\end{array}
\right.
$$

\noindent
$ t_i(n_1,v,n_2) = $

\noindent
$$
\left\{
\!\begin{array}{l@{\hspace{1mm}}l}
\! f_{so}(n_1,v,n_2) + f_{so}(n_2,v,n_1), & \mbox{if $i = 3$}\\
\! f_s(n_1,v) \!+\! f_o(n_1,v) \!+\! f_s(n_2,v) \!+\! f_o(n_2,v), & \mbox{if $i = 2$}\\
\! f_s(v) + f_o(v), & \mbox{if $i = 1$}\\
\end{array}
\right.
$$

\noindent
The definition of $P_3(1|n_1,v,n_2)$ is analogous to that of
$P_{bo}(w_n|w_1^{n-1})$. In the case where the counts are
positive, the numerator in the latter is the number of times the
word $w_n$ followed the $n$-gram $w_1^{n-1}$ in training
data, and in the former, the number of times $n_1$ occurred as
the subject with $n_2$ as the object of $v$. This count is
divided, in the latter, by the number of times the $n$-gram
$w_1^{n-1}$ was seen in training data, and in the former, by
the number of times $n_1$ was seen as the subject or object of
$v$ with $n_2$ as its object/subject respectively.

However, the definition of $P_2(1|n_1,v,n_2)$ is somewhat different;
it makes use of both the subject and object relations implicit in the
tuple.  In $P_2(1|n_1,v,n_2)$, one combines the evidence for $n_1$ as
the subject of $v$ (with any object) with that of $n_2$ as the object
of $v$ (with any subject).  

At the $P_1$ level, only the counts obtained for the verb are used in
the estimate; although for certain verbs some nouns may have definite
preferences for appearing in the subject or object position, this
information was deemed on empirical grounds not to be appropriate for
all verbs.

When the verb $v$ in a test tuple $(n_1,v,n_2)$ does not occur in any
training tuple, the default $P_0(1|n_1,v,n_2)= 1.0$ is used; it
reflects the fact that constructs in which the first noun is the
subject of the verb are more common.

\subsection{Decision Algorithm}

The decision algorithm determines for a given test tuple
$(n_1,v,n_2)$, which noun is the subject of the verb $v$.  In case one
of the nouns in the tuple is a pronoun, it does not make sense to
predict that it is subject/object of a verb based on how often it
occurred unambiguously as such in a sample text. In this case, only
the information provided by training data for the noun in the test
tuple is used. Further, in case both heads in a test tuple are
pronouns, the tuple is not considered. The algorithm is as follows.

\noindent
If $n_1$ and $n_2$ are both nouns, then $n_1$ is the subject of $v$ if
$P_3(1|n_1,v,n_2) \ge 0.5$, else its object.

\noindent
In case $n_2$ (but not $n_1$) is a pronoun,
redefine $c_i$ and $t_i$ as follows:

$$
c_i(n_1,v,n_2) = \left\{
\begin{array}{ll}
f_s(n_1,v), & \mbox{if $i = 2$}\\
f_s(v), & \mbox{if $i = 1$}\\
\end{array}
\right.
$$

$$
t_i(n_1,v,n_2) = \left\{
\begin{array}{ll}
f_s(n_1,v) + f_o(n_1,v), & \mbox{if $i = 2$}\\
f_s(v) + f_o(v), & \mbox{if $i = 1$}\\
\end{array}
\right.
$$

\noindent
and calculate $P_2(1|n_1,v,n_2)$ with these new definitions.  If
$P_2(1|n_1,v,n_2) \ge 0.5$, then $n_1$ is the subject of the verb $v$,
else its object. We proceed analogously in case $n_1$ (but not $n_2$)
is a pronoun.

\subsection{Related Work}

In \cite{CoBro95} the back-off model is used to decide PP attachment
given a tuple $(v, n_1, p, n_2)$, where $v$ is a verb, $n_1$ and $n_2$
are nouns, and $p$ a preposition such that the PP headed by $p$ may be
attached either to the verb phrase headed by $v$ or to the NP headed
by $n_1$, and $n_2$ is the head of the NP governed by $p$.

The model presented in section~\ref{rev-sec} is similar to that
in \cite{CoBro95}, however, unlike \cite{CoBro95}, who use
examples from a treebank to train their model, the procedure
described in this paper uses training data automatically obtained
from sample text. Accordingly, the model must cope with the fact
that training data is much more likely to contain errors. The
next section evaluates the decision algorithm as well as the
training data obtained by the learning procedure.

\section{Results}
\label{sec-results}

The method described in the previous section was applied to a text
corpus consisting of 5 months of the newspaper {\em Frankfurter
Allgemeine Zeitung} with approximately 15 million word-like
tokens. The learning procedure produced a total of 24,178 test tuples
and 47,547 training triples.

\subsection{Learning procedure}

In order to evaluate the data used to train the model, 1000 training
tuples were examined. Of these tuples, 127 were considered to be
(partially) incorrect based on the judgments of a single judge given
the original sentence.  Errors in training and test data may stem from
the morphology component, from the grammar specification, from the
heuristic rule, or from actual errors in the text.

\subsubsection{Subcategorization Information}

The system works without subcategorization information; it suffices
for a verb to occur with a possibly nominative and a possibly
accusative NC for it to be considered training/test data.  Lack of
subcategorization leads to errors when verbs occurring with an
(ambiguous) dative NC are mistaken for verbs which subcategorize for
an accusative nominal phrase. For instance in (\arabic{C}) below, the
verb {\em geh\"oren} (`to belong') takes, in one reading, a dative NP
as its object and a nominative NP as its subject. Since the nominal
constituent [$_{NC}$~Bill] is ambiguous with respect to case and
possibly accusative, the erroneous tuple {\em (Wagen, geh\"oren, Bill,
1)} (`car, belong, Bill') is produced for this sentence.

\begin{sentence}
(\arabic{C})
& Der Wagen & geh\"ort & Bill. \\
& the car & belongs & Bill\\
&\makebox[0mm][l]{`The car belongs to Bill.'}\\
\end{sentence}
\stepcounter{C}
\medskip

\noindent
Another source of errors is the fact that any accusative NC is
considered an object of the verb. For instance in sentence
(\arabic{C}), the verb {\em trainieren} (`to train') occurs with two
NCs. Since the NC preceding the verb is unambiguously nominative and
the one following the verb possibly accusative, the training tuple
{\em (Tennisspieler, trainieren, Jahr, 1)} (`tennis player, train,
year') is produced for this sentence, although the second NC is not an
object of the verb.

\begin{sentence}
(\arabic{C})
& Der Tennisspieler & trainiert & das ganze Jahr.\\
& the tennis player & trains & the whole year\\
\end{sentence}
\stepcounter{C}

\subsubsection{Homographs}

In sentence (\arabic{C}) below, the word {\em morgen} (`tomorrow') is
an adverb. However, its capitalized form may also be a noun, leading
in this case to the erroneous training tuple {\em (Morgen, trainieren,
Tennisspieler, 0)} (since [$_{NC}$~der Tennisspieler] is unambiguously
nominative).

\begin{sentence}
(\arabic{C})
& Morgen & trainiert & der Tennisspieler. \\
& tomorrow & trains & the tennis player \\
&\makebox[0mm][l]{`The tennis player will train tomorrow.'}\\
\end{sentence}
\stepcounter{C}

\subsubsection{Separable Prefixes}

In German, verb prefixes can be separated from the verb.  When a
finite (separable prefix) main verb occupies the second position in
the clause, its prefix takes the last position in the clause core. For
example in sentence (\arabic{C}) below, the prefix {\em zur\"uck} of
the verb {\em zur\"uckweisen} (`to reject') follows the object of the
verb and a subordinate clause with a subjunctive main verb. This
construct is not covered by the current version of the
grammar. However, due to the grammar definition, and since {\em
weisen} is also a verb (without a separable prefix) in German, [$_C$
Er weist die Kritik der Prinzessin] is still accepted as a valid
clause, leading to the erroneous training tuple $(er, weisen, Kritik,
1)$ (`he, point, criticism'). Such errors may be avoided with further
development of the grammar.

\begin{sentence}
(\arabic{C})
& Er & weist & die Kritik & der Prinzessin, & seine \\
& he & rejects & the criticism & the princess & his \\
\end{sentence}
\begin{continuation}
 \makebox[3.4mm]  & Ohren & seien & zu gro\ss, & zur\"uck. \\
                  & ears & are & too big & PRT \\[1mm]
&\makebox[0mm][l]{`He rejects the princess' criticism that his ears }\\
&\makebox[0mm][l]{ are too big.''}\\
\end{continuation}
\stepcounter{C}

\subsubsection{Constituent Heads}

The system is not always able to determine constituent heads
correctly. For instance in sentence (\arabic{C}), all words in the
name {\em Mexikanische Verband f\"ur Menschenrechte} are
capitalized. Upon encountering the adjective {\em Mexikanische}, the
system takes it to be a noun (nouns are capitalized in German),
followed by the noun {\em Verband} ``in apposition''. Sentence (11) is
the source of the erroneous training tuple {\em (Mexikanisch,
beschuldigen, Beh\"orde, 1)} (`Mexican, blame, public authorities').

\begin{sentence}
(\arabic{C}) & Der Mexikanische Verband & f\"ur Menschen-\\
             & the Mexican Association & for Human \\
\end{sentence}
\begin{continuation}
 \makebox[3.4mm] & rechte & beschuldigt & die Beh\"orden. \\
                 & Rights & blames & the public authorities\\[1mm]
&\makebox[0mm][l]{`The Mexican Association for Human Rights}\\
&\makebox[0mm][l]{ blames the public authorities.'}\\
\end{continuation}
\stepcounter{C}

\subsubsection{Multi-word lexical units}

The learning procedure has no access to multi-word lexical units. For
instance in sentence (\arabic{C}), the first word in the expression
{\em Hand in Hand} is considered the object of the verb, leading to
the training tuple {\em (Architekten, arbeiten, Hand, 1)} (`architect,
work, hand'). Given the information the system has access to, such
errors cannot be avoided.

\begin{sentenceNarrow}
(\arabic{C})
& Alle Architekten & sollen & Hand in Hand & arbeiten.\\
& all  architects & should  & hand in hand & work\\[1mm]
&\makebox[0mm][l]{`All architects should work hand in hand.'}\\
\end{sentenceNarrow}
\stepcounter{C}

\subsubsection{Source Text}

Not only spelling errors in the source text are the source of
incorrect tuples. For instance in sentence (\arabic{C}), the verb {\em
suchen} (`to seek') is erroneously in the third person plural. Since
{\em Reihe} (`series') in German is a singular noun, and {\em
Kontakte} (`contacts') plural, the actual object, but not the subject,
agrees in number with the verb, so the incorrect tuple {\em (Reihe,
suchen, Kontakt, 0)} (`series, seek, contact') is obtained from this
sentence.

\begin{starredsentence}
(\arabic{C})
& *&Eine Reihe & von Staaten & suchen & gesch\"aftliche \\
&  &a series & from states & seek & business \\
\end{starredsentence}
\begin{continuation}
\makebox[4.2mm] & Kontakte & zu der Region.\\
                & contacts & to the region \\
& \makebox[0mm][l]{`*A series of states seek contacts to the region.'}\\
\end{continuation}
\stepcounter{C}

\noindent
Finally, a large number of errors, specially in test tuples, stems from
the fact that soft constraints are used for words unknown to the
morphology.

\subsection{Decision Algorithm}

\begin{figure*}[htb]
\label{fig-tuples}
\begin{center}
\begin{tabular}{|l|c|c|c|c|} \hline
$P_n$ & Number & Percent of test tuples & Number correct & Accuracy \\ \hline
$P_3$ &   2    &  0.28    &   2 & 100.00 \\ 
$P_2$ & 204    & 28.53    & 194 & 95.10  \\ 
$P_1$ & 486    & 67.97    & 431 & 88.68  \\ 
$P_0$ &  23    &  3.22    &  20 & 86.96  \\ 
Total & 715    & 100.00   & 647 & 90.49  \\ \hline
\end{tabular}
\caption{\label{fig-accuracy}The accuracy of the system at each level}
\end{center}
\end{figure*}

In order to evaluate the accuracy of the decision algorithm, 1000
triples were selected from the set of test triples.  Of these, 285
contained errors, based on the judgements of a single judge given the
original sentence\footnote{The higher error rate for test tuples is
due to the soft constraints used for words unknown to the
morphology.}. The results produced by the system for the remaining 715
tuples were compared to the judgements of a single judge given the
original text. The system performed with an overall accuracy of
90.49\%.

A lower bound for the accuracy of the decision algorithm can be
defined by considering the first noun in every test tuple to be the
subject of the verb (by far the most common construct), yielding for
these 715 tuples an accuracy of 87.83\%.

The above figure shows how many of the 715 evaluated test tuples were
assigned subject/object based on the values $P_n$, and the accuracy of
the system at each level.

The accuracy for $P_2$ and $P_3$ exceeds 95\%. However, their coverage
is relatively low (28.81\%). Since the procedure used to collect
training data runs without supervision, increasing the size of the
training set depends only on the availability of sample text and
should be further pursued.

One reason for the relatively low coverage is the fact that German
compound nouns considerably increase the size of the sample space.
For instance, the head of the nominal constituent [$_{NC}$~Der
Tennisspieler] (`the tennis player') is considered by the system to be
the compound noun {\em Tennisspieler} (`tennis player'), instead of
its head noun {\em Spieler} (`player'). Consistently considering the
head of putative compound nouns to be the head of nominal constituents
may in some cases lead to awkward results. However, reducing the size
of the sample space by morphological processing of compound nouns
should be considered in order to increase coverage.

\subsubsection{Examples}

Following are examples of test tuples for which a decision was made
based on values of $P_2$. All sentences below stem from the corpus.

Sentence (\arabic{C}) was the source for the test tuple {\em
(Ausstellung, zeigen, Spektrum)} (`exhibition, show, spectrum'). This
tuple was correctly disambiguated with $P_2 = 0.87$, with, among
others, the training tuples {\em (Ausstellung, zeigen, Bild, 1)}
(`exhibition, show, painting'), {\em (Ausstellung, zeigen, Beispiel,
1)} (`exhibition, show, example'), and {\em (Ausstellung, zeigen,
Querschnitt, 1)} (`exhibition, show, cross-section') obtained with the
Agreement (sentences (15) and (16)) and Case Rules (sentence (17)),
respectively.

\begin{sentence}
(\arabic{C}) & Die Ausstellung & zeigt & das Spektrum & j\"udischer \\
             & the exhibition & shows & the spectrum & jewish \\
\end{sentence}
\begin{continuation}
\makebox[3.4mm] & Buchkunst & von den Anf\"angen [$\ldots$]\\
              & book art & from the beginnings \\[1mm]
&\makebox[0mm][l]{`The exhibition shows the spectrum of jewish}\\
&\makebox[0mm][l]{ book art from the beginnings [$\ldots$].'}\\
\end{continuation}
\stepcounter{C}

\begin{sentence}
(\arabic{C}) & die letzte Ausstellung & vor der Sommerpause\\
             & the last exhibition & before the summer pause\\
\end{sentence}
\begin{continuation}
\makebox[3.4mm][l] & zeigt & Bilder und Zeichnungen & von Petra \\
              & shows &  paintings und drawings & from Petra \\
\end{continuation}
\begin{continuation}
\makebox[3.4mm] & Trenkel & zum Thema ``Dorf''.\\
                & Trenkel &to the subject village\\[1mm]
&\makebox[0mm][l]{`The last exhibition before the summer pause }\\
&\makebox[0mm][l]{ shows paintings and drawings by Petra }\\
&\makebox[0mm][l]{ Trenkel on the subject ``village''.'}\\
\end{continuation}
\stepcounter{C}

\begin{sentence}
(\arabic{C}) & Die Ausstellung & im Museum & f\"ur Kunst-\\
             & the exhibition & in the museum & for arts and \\
\end{sentence}
\begin{continuation}
\makebox[3.4mm] & handwerk & zeigt & Beispiele & seiner vielf\"altigen \\
                & crafts & shows & examples & his manifold \\
\end{continuation}
\begin{continuation}
\makebox[3.4mm] & Objekt-Typen [$\ldots$]\\
              & object types\\[1mm]
&\makebox[0mm][l]{`The exhibition in the museum for arts and}\\
&\makebox[0mm][l]{ crafts shows examples of his manifold}\\
&\makebox[0mm][l]{ object types [$\ldots$]'}\\
\end{continuation}
\stepcounter{C}

\begin{sentence}
(\arabic{C}) & Eine & vom franz\"osischen Kulturinstitut \\
             & a    & from the French culture institute \\
\end{sentence}
\begin{continuation}
\makebox[3.4mm] & mit Unterst\"utzung des B\"orsenvereins \\
                & with  support the B\"orsenverein \\
\end{continuation}
\begin{continuation}
\makebox[3.4mm] & in der Zentralen Kinder- und Jugendbibliothek \\
                 & in the central children and youth library \\
\end{continuation}
\begin{continuation}
\makebox[3.4mm] & im B\"urgerhaus Bornheim  \\
                & in the community center Bornheim  \\
\end{continuation}
\begin{continuation}
\makebox[3.4mm] & eingerichtete Ausstellung & zeigt \\
                & organized exhibition& shows \\
\end{continuation}
\begin{continuation}
\makebox[3.4mm] & einen interessanten Querschnitt.\\
                & an interesting cross-section\\[1mm]
&\makebox[0mm][l]{`A exhibition in the central children's and }\\
&\makebox[0mm][l]{ youth library in the community center Born-}\\
&\makebox[0mm][l]{ heim, organized by the French culture }\\
&\makebox[0mm][l]{ institute with support of the B\"orsenverein, }\\
&\makebox[0mm][l]{ shows an interesting cross-section.'}\\
\end{continuation}
\stepcounter{C}

\smallskip
\noindent
Sentence (\arabic{C}) below was the source for the test tuple
{\em(Altersgrenze, nennen, Gesetz)} (`age limit, mention, law'). The
system incorrectly considered the noun {\em Altersgrenze} to be the
subject of the verb.

\begin{sentence}
(\arabic{C}) & Eine Altersgrenze & nennt & das Gesetz & nicht.\\
             & an age limit & mentions & the law & not \\[1mm]
&\makebox[0mm][l]{`The law does not mention an age limit.'}\\
\end{sentence}
\stepcounter{C}

\smallskip
\noindent
There were no training tuples in which the compound noun {\em
Altersgrenze} occurred as the subject/object of the verb. However, the
noun {\em Gesetz} occurred more frequently as the object of the verb
{\em nennen} than as its subject, leading to the erroneous decision.

\section{Conclusion}

This paper describes a procedure to automatically assign grammatical
subject/object relations to ambiguous German constructs. It is based
on an unsupervised learning procedure to collect test and training
data and the back-off model to make assignment decisions. The system
was implemented and tested on a 15-million word newspaper corpus.

The overall accuracy of the decision algorithm was almost 3\% higher
than the baseline of 87.83\% established. The accuracy of the
procedure for tuples for which a decision was made based on training
pairs/triples ($P_2$ and $P_3$) exceeded 95\%.

In order to increase the coverage for these cases as well as the
overall performance of the procedure, the sample space should be
reduced by morphologically processing German compound nouns, and the
size of the training set should be increased. Further, in the
experiment described in this paper, the model was trained with data
obtained by an unsupervised procedure which performs with an accuracy
of approximately 87\% for training data. Further development of the
morphology component and grammar definition should lead to improved
results.

\section{Acknowledgments}

I would like to thank Michael K\"onyves-T\'oth, who developed the
parser engine used in the experiment described in this paper, for his
support. I would also like to thank Martin B\"ottcher and the
anonymous reviewers for many helpful comments on an earlier version of
the paper.

\end{document}